\documentclass[12pt]{article}
\usepackage{amssymb}                         
\usepackage[cp1251]{inputenc}
\usepackage[russian]{babel}
\usepackage{graphicx}
\usepackage[russian]{babel}

\newtheorem{myalgorithm}{Алгоритм}

\makeatletter
\renewcommand{\@begintheorem}[2]{                                  
\begin{trivlist}\sl\item[\hspace{\labelsep}{\bf #1\ #2\,.}]}       
\renewcommand{\@opargbegintheorem}[3]{                             
\begin{trivlist}\sl\item[\hspace{\labelsep}{\bf #1\ #2\ (#3)\,.}]} %
\renewcommand{\@endtheorem}{\end{trivlist}}                        %
\makeatother
\makeatletter
\newcommand{\stylesection}
{ \setcounter{section}{0}
\renewcommand{\section}
{
  \secdef\Section\sSection
 }
}
\newcommand{\Section}[2][?]
{
 \refstepcounter{section}
 \addcontentsline{toc}{section}{\thesection.~#1}
 {
  \center\large\bfseries\thesection.
 \vspace{\baselineskip}~#2\par
 }
 \sectionmark{#1}
}
\newcommand{\sSection}[1]
{
 {
  \center\large\bfseries
  #1\par
 }
 \sectionmark{#1}
 \vspace{1\baselineskip} 
}
\stylesection 
\renewcommand{\ps@plain}{
\renewcommand{\@oddhead}{} 
\renewcommand{\@oddfoot}{\hfil\thepage\hfil} }
\renewcommand{\@makecaption}[2]{
\vspace{\abovecaptionskip}
\sbox{\@tempboxa}{#1. #2}                
\ifdim \wd\@tempboxa >\hsize #1. #2\par  
\else \global\@minipagefalse             
\hbox to \hsize {\hfil #1. #2\hfil}
\fi \vspace{\belowcaptionskip}}
\newcommand{\@maketablecaption}[2]{
\begin{flushright}#1\par\end{flushright}
}
\renewcommand{\table}{\let\@makecaption\@maketablecaption\@float{table}}

\begin{document}

\pagestyle{plain}
\title{Исследование оптимальной рекомбинации в генетическом алгоритме для одной задачи составления расписаний с переналадками}

\author{А.В.~Еремеев, Ю.В.~Коваленко}

\maketitle

\begin{abstract}
{В работе проводится экспериментальное исследование оператора
оптимальной рекомбинации для задачи минимизации общего времени
завершения работ на одном устройстве с учетом переналадок
($1|s_{vu}|C_{\max}$). Вычислительный эксперимент, проведенный на
тестовых примерах из библиотеки TSPLIB, показал целесообразность
решения задачи оптимальной рекомбинации в операторе кроссинговера
генетического алгоритма для~$1|s_{vu}|C_{\max}$.
}\\
\\
\textbf{Ключевые слова:} расписание, переналадка, генетический
алгоритм, оптимальная рекомбинация.
\end{abstract}

{
\section*{Введение}
\label{sec:Introduction} \sloppy
\par
Работоспособность генетических алгоритмов (см.,
например,~\cite{RutPilRut,Holl75,Reeves}) существенно зависит от
выбора оператора кроссинговера, где комбинируются элементы
родительских решений при построении решений-потомков. {\em Задача
оптимальной рекомбинации}~(ЗОР) состоит в отыскании наилучшего
возможного результата кроссинговера при заданных двух родительских
решениях комбинаторной задачи оптимизации. Результаты,
содержащиеся в~\cite{ErKov1,BN98,GLM00} и других работах, дают
экспериментальное подтверждение целесообразности решения ЗОР в
операторах кроссинговера. В генетических алгоритмах для задач, где
множество допустимых решений составляют перестановки, алгоритмы
кроссинговера подобного типа применялись в работах M.~Yagiura и
T.~Ibaraki~\cite{YI}, C.~Cotta, E.~Alba и J.M.~Troya~\cite{CAT},
W.~Cook и P.~Seymour~\cite{Cook} и D.~Whitley, D.~Hains и
A.~Howe~\cite{WHH10}. Особый интерес представляют NP-трудные
комбинаторные задачи оптимизации, для которых ЗОР полиномиально
разрешима. Впервые такие задачи были обнаружены E.~Balas и
W.~Niehaus в~\cite{BN98}, где был предложен эффективный оператор
кроссинговера для задачи о наибольшей клике.

В настоящей работе рассматривается следующая задача теории
расписаний. Задано множество работ $V=\{v_1,\dots,v_k\}$, которые
должны быть выполнены на единственном имеющемся устройстве. Каждая
работа $v\in V$ характеризуется длительностью $p_v\in {\mathbb
{R}}^+$ (здесь ${\mathbb {R}}^+$~--~множество положительных
вещественных чисел). Прерывание выполнения работ не допускается. В
каждый момент времени устройство не может быть задействовано более
чем в одной работе. При этом если устройство переключается с одной
работы на другую, то необходимо выполнять переналадку. Пусть
$s_{vu}\in {\mathbb {R}}_+$~--~длительность переналадки с
работы~$v$ на работу~$u$ для всех $v,u\in V$, где $v\ne u$ (здесь
и далее ${\mathbb {R}}_+$~--~множество неотрицательных
вещественных чисел). Требуется составить расписание,
минимизирующее время завершения всех работ.

Обозначим через $\pi=(\pi_1,\dots,\pi_k)$ перестановку,
определяющую порядок выполнения работ, а именно,
$\pi_i$~--~работа, выполняемая $i$-той по счету. Пусть
$s(\pi)=\sum\limits_{i=1}^{k-1}s_{\pi_i,\pi_{i+1}}$. 
Тогда задача эквивалентна поиску такой перестановки $\pi^*$, при
которой минимизируется суммарная длительность переналадки $
s(\pi^*)$.

В соответствии с известными обозначениями~\cite{GLLR} данная
задача обозначается через $1|s_{vu}|C_{\max}$, и является
NP-трудной в сильном смысле, так как к ней полиномиально сводится
NP-полная в сильном смысле задача ГАМИЛЬТОНОВ ПУТЬ~\cite{GJ}.

В работе~\cite{ErKo12} исследуется вычислительная сложность задачи
поиска для произвольных заданных {\it родительских} решений
$\pi^1$ и $\pi^2$ такой перестановки $\pi'$, при которой\\
1)~$\pi_i'=\pi_i^1$ или $\pi_i'=\pi_i^2$ для всех $i=1,\dots,k$;\\
2)~$\pi'$ имеет минимальное значение целевой функции $s(\pi')$
среди всех перестановок, удовлетворяющих условию~1).

Поиск такой перестановки представляет собой задачу оптимальной
рекомбинации для задачи~$1|s_{vu}|C_{\max}$.

В~\cite{ErKo12} с использованием результатов
А.И.~Сердюкова~\cite{SAI} показана NP-трудность в сильном смысле
представленной задачи оптимальной рекомбинации, предложен алгоритм
решения исследуемой задачи, основанный на переборе совершенных
паросочетаний в специальном двудольном графе и доказано, что для
<<почти всех>> пар родительских решений данный алгоритм имеет
полиномиальную трудоемкость.

Настоящая статья построена следующим образом. В
$\S$~\ref{sec:GAforOR} описывается генетический алгоритм с
элитарной рекомбинацией для рассматриваемой задачи. В
$\S$~\ref{sec:solution} приводится описание оператора оптимальной
рекомбинации, предложенного в~\cite{ErKo12};
$\S$~\ref{sec:experimentOR} содержит результаты вычислительного
эксперимента. В последнем параграфе содержатся заключительные
замечания.

\section{Генетический алгоритм с элитарной рекомбинацией}
\label{sec:GAforOR}

Генетический алгоритм является алгоритмом случайного поиска, в
котором моделируется процесс развития популяции особей,
соответствующих решениям задачи оптимизации. Новые особи-потомки
строятся на основе имеющихся особей путем применения вероятностных
операторов кроссинговера и мутации. Пусть
$\Pi^t=(\zeta^{1,t},\dots,\zeta^{r,t})$ обозначает популяцию
поколения~$t,$ ${t=0,1,2,\ldots}$, где $r$~--~численность
популяции, которая является фиксированной от начала работы
алгоритма и до конца.

В~\cite{GoldTheir94} описывается ГА с {\it элитарной
рекомбинацией} и полной заменой популяции, который заключается в
следующем. Особи текущей популяции~$\Pi^t$ случайным образом
переставляются и выбираются пары родительских особей в порядке
возрастания номеров $(\zeta^{1,t}, \zeta^{2,t}), (\zeta^{3,t},
\zeta^{4,t}), ...$. К каждой из выбранных пар применяется оператор
кроссинговера, результатом которого являются две особи. Пара
потомков сравнивается с соответствующими родительскими особями, и
лучшие две из четырех особей помещаются в новую популяцию. Такой
способ обновления популяции впервые был предложен D.~Cavicchio
в~\cite{Cavicchio1970}, где он был называн {\it предварительной
селекцией} (preselection).

Для представления решений задачи~$1|s_{vu}|C_{\max}$ достаточно
естественно использовать перестановки. В работе M.~Yagiura,
T.~Ibaraki~\cite{YI} для решения задач оптимизации, где множество
допустимых решений составляют перестановки, в том числе для задачи
коммивояжера, предлагается использовать генетический алгоритм,
представляющий собой ГА с {элитарной рекомбинацией} при
{стационарной схеме воспроизведения} (см.,
например,~\cite{Reeves}). Отличие данного ГА от
алгоритма~\cite{GoldTheir94} заключается в том, что на каждой
итерации обновляется не более двух особей в популяции. Для этого
в~\cite{YI} случайным образом выбирается только одна пара
родительских особей $p^1$ и $p^2$, которая сравнивается со своими
потомками, и лучшие две из этих особей замещают  $p^1$ и $p^2$.

Опишем ГА с {элитарной рекомбинацией} при {стационарной схеме
воспроизведения} для случая, когда в качестве оператора
кроссинговера используется оптимальная рекомбинация. Для краткости
изложения такой алгоритм будем называть {\it ГА с элитарной
оптимальной рекомбинацией}.

\begin{myalgorithm} {\bf ГА с элитарной оптимальной
рекомбинацией}\label{algorithm:GA_elitist_recombination}\end{myalgorithm}

{\sc Шаг~0}.~Положить $t:=0$.

{\sc Шаг~1}.~Построить начальную популяцию $\Pi^0$.

{\sc Шаг~2}.~Пока не выполнен критерий остановки, выполнять:

\mbox{\hspace{1em}} {\sc 2.1.}~Выбрать  из популяции $\Pi^t$ две
родительских особи~$p^1, p^2$ случайным образом с равномерным
распределением.

\mbox{\hspace{1em}} {\sc 2.2.}~Применить к $p^1$ и $p^2$ оператор
{мутации}.

\mbox{\hspace{1em}} {\sc 2.3.}~Построить потомка $p'$, применяя
оператор оптимальной рекомбинации к~$p^1, p^2$.

\mbox{\hspace{1em}} {\sc 2.4.}~Заменить потомком~$p'$ одного из
двух родителей. Положить ${\Pi^{t+1}:=\Pi^t}$.

\mbox{\hspace{1em}} {\sc 2.5.}~Положить $t:=t+1$.

{\sc Шаг~3}.~Результатом работы алгоритма является лучшее
найденное решение.\\

В предлагаемом ГА с элитарной оптимальной рекомбинацией для
решения задачи~$1|s_{vu}|C_{\max}$ особи, как и допустимые
решения, представляются перестановками, а на шаге~2.3 применяется
оператор оптимальной рекомбинации~1)~--~2). Определим метод
построения начальной популяции, способ замены родительских особей
потомком и оператор мутации.


{\bf Построение начальной популяции.}\   Особи начальной популяции
строятся методом {\it случайного присоединения} (arbitrary
insertion)~\cite{YI}. Это жадный алгоритм, который позволяет
генерировать достаточно разнообразные решения приемлемого качества
за короткое время.

Перестановку подмножества работ $V'\subseteq V$ будем называть
частичным расписанием. Метод случайного присоединения стартует со
случайно выбранного частичного расписания, содержащего только две
работы, и состоит из $|V|-2$ шагов. Каждый шаг заключается в
выборе работы~$u$ случайным образом из числа еще неразмещенных и
ее вставке между парой соседних работ~$v_i$~и~$v_j$ в текущем
частичном расписании. Соседние работы~$v_i$~и~$v_j$ выбираются
таким образом, чтобы величина ${s_{v_i,u}+s_{u, v_j}-s_{v_i,
v_j}}$ была минимальна. В результате будет построена перестановка,
представляющая допустимое решение задачи~$1|s_{vu}|C_{\max}$.

{\bf Замена родительских особей потомком.} \ Предположим, что
$s(p^1)\leqslant s(p^2)$. Если $s(p')< s(p^1)$, то потомок~$p'$
замещает родителя $p^2$, в противном случае потомок не добавляется
в популяцию. Заметим, что с ростом числа итераций $p^1$ и $p'$
становятся достаточно близкими друг к другу, что может
способствовать быстрому снижению разнообразия популяции в ГА. Для
предотвращения этого недостатка в~\cite{YI} предлагается
модифицировать указанный метод следующим образом. С вероятностью
$P(\Delta_1/\Delta_2)$ потомок $p'$ замещает $p^2$, иначе $p'$
замещает $p^1$, где
\begin{equation}
\Delta_i=s(p^i)-s(p'),\ i=1,2,
\end{equation}
\begin{equation}
P(\Delta_1/\Delta_2)=\min\Big\{\frac{\Delta_1/\Delta_2}{a},1\Big\}.
\end{equation}

Константа $a\geqslant 0$ является настраиваемым параметром ГА.
Отметим, что $\Delta_2\geqslant \Delta_1\geqslant 0$ по
определению и, следовательно, $\Delta_1/\Delta_2\in [0,1]$
(предполагается, что $\Delta_1/\Delta_2=1$, если
$\Delta_1=\Delta_2=0$). Ясно, что при $a=0$ потомок $p'$ всегда
будет замещать $p^2$, а при $a=\infty$ $p'$ всегда будет замещать
$p^1$.

{\bf Оператор мутации.}  \ В качестве оператора мутации можно
использовать, например, мутацию {\it сдвига} (shift mutation) или
{\it обмена} (exchange mutation). В~\cite{YI} экспериментальным
путем установлено, что ГА с элитарной рекомбинацией показывает
достаточно хорошие результаты и без применения мутации
к родительским особям. 
Это связано с тем, что оператор мутации используется для того,
чтобы избежать бытрого <<сужения>> области поиска в
ГА~\cite{RutPilRut}. Однако описанный способ замены родительских
особей потомком, как правило, предотвращает преждевременную
сходимость
ГА и без оператора мутации. 

\section{Решение задачи оптимальной рекомбинации} \label{sec:solution}
\sloppy
\par
{Рассмотрим вспомогательную задачу~${I}$ о {\em кратчайшем
гамильтоновом пути с предписаниями вершин} следующего вида.} Задан
полный ориентированный граф~$G=(X,U)$, где
${X=\{x_1,\dots,x_n\}}$~--~множество вершин, $U=\{(x,y):\ x,y \in
X, x\ne y\}$~--~множество дуг, которым приписаны веса
${\rho(x,y)\in {\mathbb {R}}_+,}$ ${(x,y)\in U}$. {Также задана
система  подмножеств (предписаний) $X^i\subseteq X,\ i=1,\dots,n,$
удовлетворяющая
условиям:}\\
$\mathcal{C}1$:~$|X^i|\leqslant 2$ для всех ${i=1,\dots,n}$;\\
$\mathcal{C}2$:~${1 \leqslant|\{i:\ x\in X^i,\ i=1,\dots,n\}|\leqslant 2}$ для всех ${x\in X}$;\\
$\mathcal{C}3$:~если ${x\in X^i}$ и ${x\in X^j}$, где ${i\ne j}$,
то ${|X^i|=|X^j|=2}$, а если ${x\in X^i}$ только при одном
значении~$i$, то ${|X^i|=1}$.

Обозначим через $F$ множество взаимно однозначных отображений из
$X_n=\{1,\dots,n\}$ в $X$ таких, что $f(i)\in X^i,\ i=1,\dots,n,$
для любого $f\in F$. Задача~${I}$ состоит в выборе такого
отображения ${f}^*\in F$, что ${\rho}({f}^*)=\min\limits_{f\in F}
{\rho}(f)$, где ${\rho}({f})=\sum\limits_{i=1}^{n-1}
\rho({f}(i),{f}(i+1))$ для всех $f\in F$.

Если работе $v_i\in V$ поставить в соответствие вершину $x_i\in X$
графа $G$, ${i=1,\dots,k}$; положить число вершин $n=k$; веса дуг
$\rho(x,y)=s_{xy}$ для всех $x,y\in X$, где $x\ne y$; предписания
$X^i=\{\pi^1_i,\pi_i^2\}$, $i=1,\dots,k$, то множество допустимых
решений задачи $I$ будет взаимно однозначно отображаться на
множество допустимых решений задачи оптимальной рекомбинации
1)~--~2), причем оптимальному решению будет соответствовать
оптимальное.

Оптимальное отображение $f^*\in F$ в задаче~$I$ можно найти за
время $O(2^k)$ путем перебора решений (среди которых будут как
допустимые, так и недопустимые), формируемых выбором в каждом
подмножестве системы предписаний одного из его элементов. Такую же
трудоемкость имеет очевидная модификация алгоритма M.~Held и
R.M.~Karp~\cite{HK}, известного для задачи коммивояжера. Однако
можно построить более эффективный алгоритм для решения задачи~$I$,
используя подход А.И.~Сердюкова~\cite{SAI}.

Рассмотрим двудольный граф~$\bar{G}=(X_{k},X,\bar{U})$ с равными
по мощности долями вершин $X_k,\ X$  и множеством ребер
$\bar{U}=\{(i,x):$ ${i\in X_{k},\ x\in X^i}\}$. Заметим, что между
множеством допустимых решений $F$ задачи~$I$ и множеством
совершенных паросочетаний $\mathcal{W}$ в графе~$\bar{G}$
существует взаимно однозначное соответствие.

Ребро~$(i,x)\in \bar{U}$ назовем особым, если $(i,x)$ принадлежит
любому совершенному паросочетанию в графе~$\bar{G}$. Помеченными
будем называть вершины графа~$\bar{G}$, инциндентные особым
ребрам. {Под блоком в графе~$\bar{G}$ будем понимать максимальный}
{связный подграф, содержащий не менее двух ребер.} Заметим, что в
каждом блоке~$j$ графа~$\bar{G}$ степень любой вершины равна двум,
$j=1,\dots,q(\bar{G})$, где $q(\bar{G})$~--~число блоков в
графе~$\bar{G}$. Тогда ребра $(i,x)\in \bar{U}$ такие, что
$|X^i|=1$, являются особыми и блокам не принадлежат, а ребра
$(i,x)\in \bar{U}$ такие, что $|X^i|=2$, содержатся в блоках.
Кроме того, каждый блок ${j,\ j=1,\dots,q(\bar{G})}$,
графа~$\bar{G}$ имеет ровно два максимальных (совершенных)
паросочетания, наборы ребер в которых различны, поэтому, не
содержит особых ребер. Следовательно, ребро~$(i,x)\in \bar{U}$
является особым тогда и только тогда, когда $|X^i|=1$, и любое
совершенное паросочетание в графе~$\bar{G}$ взаимно однозначно
определяется набором максимальных паросочетаний (по одному из
каждого блока) и совокупностью особых ребер.

Для примера рассмотрим задачу~$I$, в которой $n=k=7$ и система
предписаний имеет вид: $X^{1}=\{x_3,x_7\}$, $X^{2}=\{x_3,x_7\}$,
$X^{3}=\{x_2\}$, $X^{4}=\{x_5\}$, $X^{5}=\{x_1,x_4\}$,
$X^{6}=\{x_4,x_6\}$, ${X^{7}=\{x_1,x_6\}}$. Соответствующий данной
задаче двудольный граф $\bar{G}=(X_7,X,\bar{U})$, его особые ребра
и блоки представлены на рис.~\ref{fig1}. Здесь ребра каждого
блока, выделенные жирным, образуют одно максимальное паросочетание
блока, а остальные ребра блока~--~второе.

\begin{figure}[!h]
\begin{center}
\vspace{2em}
\includegraphics[height=7.5cm,width=6cm]{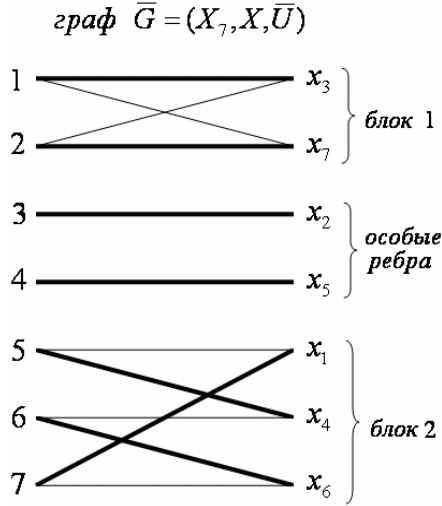}\\
\caption{Пример графа $\bar{G}=(X_7,X,\bar{U})$, содержащего два
особых ребра и два блока.} \label{fig1}
\end{center}
\end{figure}

{Особые ребра и блоки в графе~$\bar{G}$ могут быть вычислены за
время $O(k)$, например,} с помощью алгоритма <<поиск в глубину>>
или <<поиск в ширину>>~\cite{KLR}. Максимальные паросочетания в
блоках {находятся очевидным образом за время $O(k)$}.

Таким образом, для решения задачи~$I$ можно предложить следующий
алгоритм. Строим двудольный граф~$\bar{G}$, определяем в нем набор
особых ребер и блоков, а также находим максимальные паросочетания
в блоках. Перебираем все совершенные паросочетания $W\in
\mathcal{W}$ в графе~$\bar{G}$ (формируя их из максимальных
паросочетаний в блоках и особых ребер). Каждому $W\in \mathcal{W}$
ставим в соответствие $f\in F$ и вычисляем $\rho(f)$. В результате
находим $f^*\in F$ такое, что $\rho(f^*)=\min\limits_{f\in
F}\rho(f)$.

Поскольку $|F|=|\mathcal{W}|=2^{q(\bar{G})}$, то вычислительная
сложность представленного алгоритма равна $O(k\cdot
2^{q(\bar{G})})$,
причем ${q(\bar{G})\leqslant \lfloor 
\frac{k}{2}\rfloor}$ и данная оценка достижима. Рассмотрим
модификацию данного алгоритма, трудоемкость которой равна
$O(q(\bar{G})\cdot 2^{q(\bar{G})})$.

Прежде, чем приступить к перебору всех возможных комбинаций максимальных
паросочетаний в блоках, проведем предварительные подсчеты, которые позволят
ускорить вычисления значений целевой функции в процессе перебора. Условимся
называть контактом между блоком~$j$ и блоком $j'\ne j$ (или особым ребром)
пару вершин~$(i,i+1)$ левой доли графа~$\bar{G}$, причем одна из этих двух
вершин принадлежит блоку~$j$, а другая -- блоку~$j'$ (или особому ребру).
Контактом внутри блока будем называть пару вершин левой доли блока, если их
порядковые номера отличаются на единицу. Для каждого блока ${j,\
j=1,\dots,q(\bar{G}),}$ установим наличие контактов внутри блока~$j$, между
блоком~$j$ и всеми особыми ребрами, а также между блоком~$j$ и каждым другим
блоком.
Трудоемкость проверки всех вершин в левой доле одного блока на наличие
контактов есть $O(k)$.

Каждое из двух максимальных паросочетаний $w^{0,j}$ и $w^{1,j}$ блока~${j}$
определяет некоторую дугу графа~$G$ для любого контакта $(i,i+1)$ внутри
данного блока и для любого контакта блока $j$ с особыми ребрами.
Просуммируем веса дуг графа~$G$ по контактам внутри этого блока и по всем
контактам блока~$j$ с особыми ребрами, и обозначим полученные величины для
паросочетаний $w^{0,j}$ и $w^{1,j}$, соответственно, через $P^{0}_j$ и
$P^1_j$. Если блок~${j}$ контактирует с блоком~${j'\ne j}$, то каждая
комбинация максимальных паросочетаний данных блоков определяет некоторую
дугу графа~$G$ для любого контакта $(i,i+1)$ между этими блоками.
Просуммируем веса  дуг графа~$G$ по всем контактам между блоками~${j}$ и
${j'}$, и обозначим четыре полученные величины через $P^{0-0}_{j j'}$,
$P^{0-1}_{j j'}$, $P^{1-0}_{j j'}$ и $P^{1-1}_{j j'}$.

Вышеуказанные величины вычисляются для каждого блока, поэтому
общая трудоемкость предварительной процедуры есть $O(k\cdot
q(\bar{G}))$.

После этого перебор всех возможных комбинаций максимальных
паросочетаний в блоках осуществляется с помощью кода Грея (см.,
например,~\cite{RND}) таким образом, что каждая следующая
комбинация отличается от предыдущей заменой максимального
паросочетания только в одном из блоков. Пусть двоичный вектор
$\delta=(\delta_1,\dots,\delta_{q(\bar{G})})$ определяет
назначение максимальных паросочетаний в блоках, а именно,
$\delta_j=0$, если в блоке~$j$ выбрано паросочетание~$w^{0,j}$, и
$\delta_j=1$, если в блоке~$j$ выбрано паросочетание~$w^{1,j}$.
Таким образом, каждому вектору~$\delta$ взаимно однозначно
сопоставляется допустимое решение~$f_{\delta}$ задачи~$I$.

Если осуществляется переход от вектора $\bar{\delta}$ к вектору
$\delta$, при котором изменяется паросочетание в блоке~$j$, то
значение целевой функции $\rho(f_{\delta})$ вычисляется через
значение целевой функции $\rho(f_{\bar{\delta}})$ по формуле
$\rho(f_{\delta})=\rho(f_{\bar{\delta}})-P^{{\bar
\delta}_j}_j+P^{{\delta}_j}_j-\sum\limits_{j'\in A(j)}P^{{\bar
\delta}_j-{\bar \delta}_{j'}}_{jj'}+\sum\limits_{j'\in
A(j)}P^{{\delta}_j-{\bar \delta}_{j'}}_{jj'}$, где
$A(j)$~--~множество блоков, контактирующих с блоком~$j$. Поскольку
$|A(j)|\leqslant q(\bar{G})$, то пересчет целевой функции требует
времени $O(q(\bar{G}))$, и общая трудоемкость представленной
модификации алгоритма решения задачи~$I$
равна~$O(q(\bar{G})\cdot2^{q(\bar{G})})$.

Таким образом, задача оптимальной рекомбинации 1)~--~2), как и
задача~$I$, может быть решена за время
$O(q(\bar{G})\cdot2^{q(\bar{G})})$. Как показано в~\cite{ErKo12},
для <<почти всех>> пар родительских решений выполняется
неравенство ${q(\bar{G})}\leqslant 1.1\cdot{\rm ln}(k)$, т.~е.
<<почти все>> индивидуальные задачи оптимальной рекомбинации
1)~--~2) могут быть решены за
время $O(k\cdot {\rm ln}(k))$, и, кроме того, имеют не более чем 
$k$ допустимых решений.

\section{Вычислительный эксперимент}
\label{sec:experimentOR}

\hspace{6mm} Для оценки возможностей предложенного оператора
оптимальной рекомбинации, главным образом, для исследования
динамики изменения размерности области его допустимых решений  с
ростом числа итераций ГА, был проведен вычислительный эксперимент
на тестовых примерах из библиотеки TSPLIB
(http://www.iwr.uni-heidelberg.de/groups/comopt/software/TSPLIB95/).
Были выбраны тестовые примеры серии ftv: ftv35 ($k=36$), ftv55
($k=56$), ftv64 ($k=65$), ftv170 ($k=171$); серии rbg: rbg323
($k=323$), rbg358 ($k=358$), rbg403 ($k=403$) и rbg443 ($k=443$) и
задача kro124p ($k=100$).


ГА с элитарной оптимальной рекомбинацией был реализован на языке
Java в среде NetBeans IDE~7.0.1. Для решения задачи оптимальной
рекомбинации использовался алгоритм из $\S$~\ref{sec:solution}.
Тестирование проводилось на ЭВМ
Intel~Core~2~Duo~CPU~E7200~2.54~ГГц, оперативная память~2~Гб.

Напомним, что задача~$1|s_{vu}|C_{\max}$ эквивалентна задаче
поиска такой перестановки~$\pi^*$, при которой суммарная
длительность
переналадок~$s(\pi^*)=\sum_{i=1}^{k-1}s_{{\pi^*_i},{\pi^*_{i+1}}}$
минимальна, причем $C_{\max}^*=s(\pi^*)+\sum_{v\in V}p_v$. Другими
словами, задача~$1|s_{vu}|C_{\max}$ эквивалентна задаче о
кратчайшем гамильтоновом пути в графе, где число вершин ${n=k}$,
каждая вершина $i$ соответствует работе $v_i$, ${i=1,\dots,n}$, а
длины дуг $c_{ij}=s_{v_i v_j}$ при $i\ne j$. Поэтому далее под
значением целевой функции решения задачи~$1|s_{vu}|C_{\max}$ будем
понимать величину $s(\pi)$, где $\pi$ -- перестановка,
соответствующая этому решению.

\textbf{Оптимальные решения задач.} Поиск оптимального решения в
задаче о кратчайшем гамильтоновом пути может быть осуществлен с
помощью модификации метода {\it отсекающих
плоскостей}~\cite{DFJ1954}, известного для задачи коммивояжера.

Задача коммивояжера, по существу, является модификацией задачи о
назначениях~\cite{PapSt1985} с дополнительным условием,
гарантирующим исключение из оптимального решения частичных циклов
(подциклов). Идея применения метода отсекающих
плоскостей~\cite{DFJ1954} к задаче коммивояжера заключается в том,
что изначально решается соответствующая задача о назначениях. Если
полученное решение содержит частичные циклы, то вводятся
дополнительные ограничения, гарантированно их исключающие. Процесс
повторяется до тех пор, пока не будет получено решение, не
содержащее частичные циклы.

Для применения метода отсекающих плоскостей к решению задачи о
кратчайшем гамильтоновом пути, вместо задачи о назначениях будем
использовать следующую задачу булевого линейного
программирования. Определим булевы переменные:\\
 {
\begin{tabular}{cll}
 $x_{ij}=$ &
 $ \left\{
\begin{tabular}{l}
$1,$ \mbox{ если из
вершины $i$ осуществляется переход в вершину $j$,}\\
$0$\mbox{ иначе;}\\
\end{tabular} \right.
$
\end{tabular}
}\\
 {
\begin{tabular}{cll}
 $y_{ij}=$ &
 $ \left\{
\begin{tabular}{l}
$1,$ \mbox{ если вершина $i$ является последней в пути,}\\
\mbox{\ \ \ \ \ а вершина $j$ -- первой,}\\
$0$\mbox{ иначе;}\\
\end{tabular} \right.
$
\end{tabular}
}\\
где $i, j= 1,\dots,n.$
%

Целевая функция и ограничения задачи имеют вид:
\begin{equation}\label{AssignmentProblem:obj}
\sum_{i=1}^n\sum_{j=1}^n{x_{ij}c_{ij}}-\sum_{i=1}^n\sum_{j=1}^n{y_{ij}c_{ij}}
\to \min,
\end{equation}
\begin{equation}\label{AssignmentProblem:rowsum}
\sum_{j=1}^n{x_{ij}}=1,\ i=1,\dots,n,
\end{equation}
\begin{equation}\label{AssignmentProblem:colsum}
\sum_{i=1}^n{x_{ij}}=1,\ j=1,\dots,n,
\end{equation}
\begin{equation}\label{AssignmentProblem:rowcolsum}
\sum_{i=1}^n\sum_{j=1}^n{y_{ij}}=1,
\end{equation}
\begin{equation}\label{AssignmentProblem:relation_x_y}
x_{ij} \geqslant y_{ij},\ i,j=1,\dots,n,
\end{equation}
\begin{equation}\label{AssignmentProblem:exclude_diagonal}
x_{ii} =0,\  y_{ii}=0,\ i=1,\dots,n,
\end{equation}
\begin{equation}\label{AssignmentProblem:domain}
x_{ij}\in \{0,1\},\ y_{ij}\in \{0,1\},\ i,j=1,\dots,n.
\end{equation}

Ясно, что если назначение переменных $x_{ij}$, согласующееся с
представленными ограничениями, задает гамильтонов цикл, то
назначение переменных $y_{ij}$ показывает какая из дуг этого цикла
исключается для получения гамильтонова пути, а значение целевой
функции~(\ref{AssignmentProblem:obj}) определяет суммарную длину
этого пути.

Оптимальные решения во всех тестовых примерах были найдены с
помощью описанного метода. В табл.~\ref{tabserHP} представлены
оптимальные значения целевой функции $s(\pi^*)$ и время их
получения ${\rm t}^*$ (в сек.) в системе моделирования GAMS~23.2.
Для решения задач булевого линейного программирования
использовался пакет~CPLEX~12.1.

\begin{table}[!h]
\begin{center}
\caption{\small{Оптимальные решения}}
\vspace{0.3cm} \label{tabserHP}
\begin{tabular}{|c|c| c| c| c|c| c| c| c|c|}
\hline
{задача} & ftv35 & ftv55 & ftv64 & kro124p & ftv170 & rbg323 & rbg358 & rbg403 & rbg443  \\
\hline $s(\pi^*)$     & $1323$  & $1485$    & $1656$  & $35227$   &$2642$     & $1299$    & $1130$     & $2432$   & $2687$\\
\hline ${\rm t}^*$ & $49$   & $104$     & $81$   & $346$     &$730$     & $3869$    & $1820$     & $3132$   & $2062$\\
\hline
\end{tabular}
\end{center}
\end{table}

\textbf{Результаты работы генетического алгоритма.} ГА с элитарной
оптимальной рекомбинацией запускался для каждой задачи $1000$ раз,
и каждый запуск выполнялся в течении $4000$ итераций для задач
серии ftv и задачи kro124p, и в течении $8000$ итераций для задач
серии rbg. Размер $r$ популяции полагался равным $30$, а значение
параметра $a$ -- равным $0.5$, как и в~\cite{YI}.

Результаты вычислительного эксперимента представлены в
табл.~\ref{tabGAforHP}. Здесь $N_{\mbox{\scriptsize{опт}}}$ --
число запусков ГА, на которых было получено оптимальное решение,
$t_{\mbox{\scriptsize{ср}}}$ -- среднее время выполнения одного
запуска ГА в сек., а
$t_{\mbox{\scriptsize{ср}}}^{\mbox{\scriptsize{опт}}}
:=t_{\mbox{\scriptsize{ср}}}\cdot
1000/{N_{\mbox{\scriptsize{опт}}}}$ -- оценка среднего времени (в
сек.) до первого получения оптимального решения при неограниченном
числе запусков ГА с указанным числом итераций.


\begin{table}[!h]
\begin{center}
\caption{\small{Результаты работы ГА с элитарной оптимальной рекомбинацией}}
\vspace{0.3cm} \label{tabGAforHP}
\begin{tabular}{|c|c| c| c| c|c| c| c| c|c|}
\hline
{задача}        & ftv35  & ftv55    & ftv64  & kro124p & ftv170 & rbg323 & rbg358 & rbg403 & rbg443  \\
\hline

$N_{\mbox{\scriptsize{опт}}}$       & $625$  & $584$    & $563$  & $815$   &$516$   & $413$    & $392$     & $342$   & $317$\\
\hline

$t_{\mbox{\scriptsize{ср}}}$       & $0.695$  & $0.701$    & $0.719$  & $2.197$   &$1.331$   & $1.631$   & $1.768$    & $1.745$  & $1.748$\\
\hline

$t_{\mbox{\scriptsize{ср}}}^{\mbox{\scriptsize{опт}}}$       & $1.11$  & $1.2$    & $1.28$  & $2.7$   &$2.58$   & $3.95$   & $4.5$    & $5.1$  & $5.5$\\

%
\hline
\end{tabular}
\end{center}
\end{table}

Из табл.~\ref{tabGAforHP} видно, что на серии~ftv ГА с элитарной
оптимальной рекомбинацией находит оптимальное решение более чем в
$50\%$ запусков, а на серии rbg -- более чем в $30\%$ запусков.
При решении задачи~kro124p ГА с элитарной оптимальной
рекомбинацией находит оптимальное решение в  $815$ случаях из
$1000$. Кроме того, на всех тестовых примерах оценка среднего
времени ГА до первого получения оптимума значительно меньше
времени нахождения оптимального решения с помощью метода
отсекающих плоскостей (см. табл.~\ref{tabserHP}).
%

Таким образом, можно сделать вывод о перспективности применения ГА
с элитарной оптимальной рекомбинацией для решения
задачи~$1|s_{vu}|C_{\max}$, и целесообразности использования
механизма перезапуска (см., например,~\cite{BDA}).

\textbf{Размерность области допустимых решений в задаче
оптимальной рекомбинации.}
 Число допустимых решений в индивидуальной задаче
оптимальной рекомбинации равно $2^{q}$, где $q:=q({\bar G})$ --
число блоков в соответствующем двудольном графе~${\bar G}$. Далее
для компактности изложения будем говорить, что индивидуальная
задача оптимальной рекомбинации содержит $q$ блоков. В
п.~\ref{sec:solution} доказано, что если выбрать пару родительских
решений с равномерным распределением, то с вероятностью,
стремящейся к~$1$ при $k\to \infty$, задача оптимальной
рекомбинации содержит не более $(1+\varepsilon){\rm ln}(k)$ блоков
при любом $\varepsilon\in(0,{\rm log}_2(e)-1]$ и является
<<хорошей>>, т.~е. полиномиально разрешимой. Возникает вопрос:
какие задачи оптимальной рекомбинации возникают в ГА с элитарной
оптимальной рекомбинацией и какова динамика их размерности с
ростом числа итераций? Описанию результатов экспериментального
исследования этого вопроса посвящен настоящий раздел.

\begin{figure}[!h]
\begin{center}
\includegraphics[width=16.0cm,height=6.0cm]{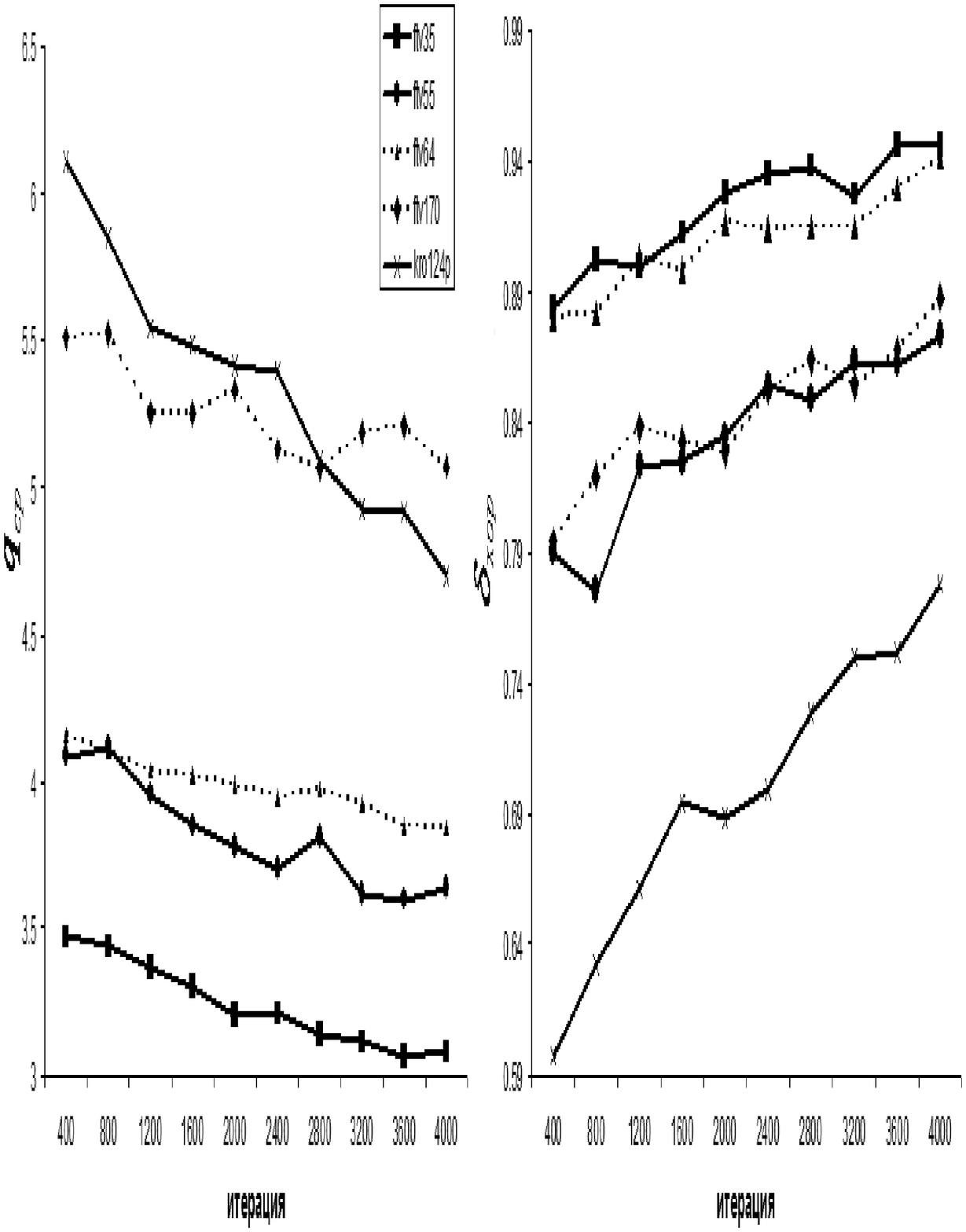}
\end{center}
\caption{Результаты для серии ftv и задачи kro124p}
 \label{ser_ftv_resOR}
\end{figure}

В ходе решения тестового примера посредством ГА с элитарной
оптимальной рекомбинацией на каждом из $1000$ запусков
запоминалось число блоков в задачах оптимальной рекомбинации с
периодичностью в $400$ итераций для задач серии ftv и задачи
kro124p, и с периодичностью в $800$ итераций для задач серии rbg.
Затем для рассматриваемых итераций вычислялось среднее число
блоков $q_{\mbox{\scriptsize{ср}}}$ и доля <<хороших>> задач
оптимальной рекомбинации $\delta_{\mbox{\scriptsize{хор}}}$ при
$\varepsilon={\rm log}_2(e)-1$ по всем запускам ГА. Полученные
значения величин  $q_{\mbox{\scriptsize{ср}}}$ и
$\delta_{\mbox{\scriptsize{хор}}}$ представлены в графической
форме на рис.~\ref{ser_ftv_resOR} и \ref{ser_rbg_resOR}.

\begin{figure}[!h]
\begin{center}
\includegraphics[width=16.0cm,height=6.0cm]{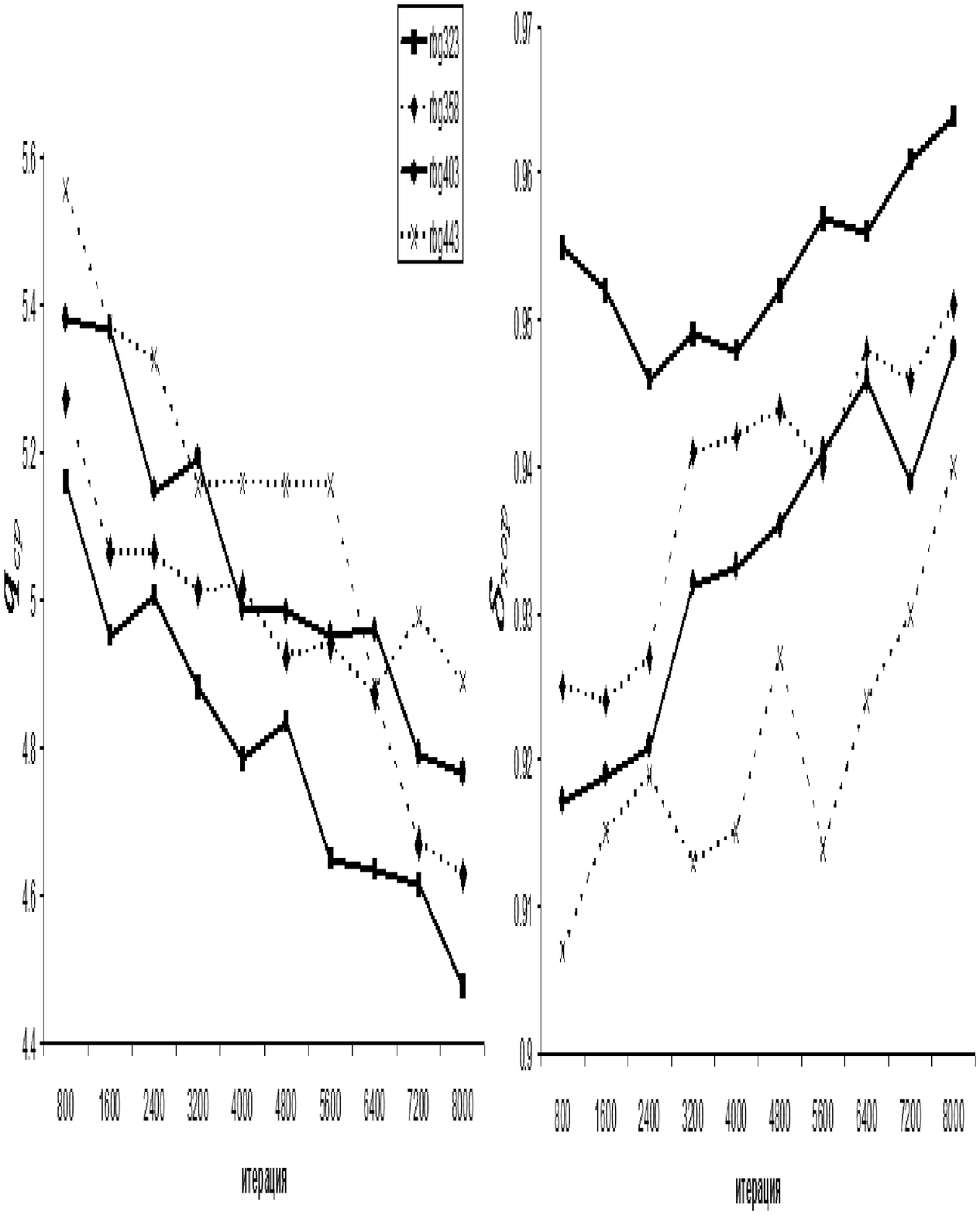}
\end{center}
\caption{Результаты для серии rbg}
 \label{ser_rbg_resOR}
\end{figure}

Видно, что на всех тестовых примерах с ростом числа итераций
среднее число блоков~$q_{\mbox{\scriptsize{ср}}}$ уменьшается.
Обозначим через  ${q_{\mbox{\scriptsize{пред}}}:={\rm ln}(k)/{\rm
ln}(2)}$ предельное значение числа блоков, при котором задача
оптимальной рекомбинации является полиномиально разрешимой.
Величина ${q_{\mbox{\scriptsize{пред}}}=5}$ для задач ftv35 и
ftv55, ${q_{\mbox{\scriptsize{пред}}}=6}$ для ftv64 и kro124p,
${q_{\mbox{\scriptsize{пред}}}=7}$ для ftv170, и
${q_{\mbox{\scriptsize{пред}}}=8}$ для всех задач серии rbg.
Обратим внимание на то, что ${q_{\mbox{\scriptsize{ср}}} \leqslant
q_{\mbox{\scriptsize{пред}}}}$ для всех тестовых примеров на всех
рассматриваемых итерациях. Кроме того, на серии ftv более чем в
$79\%$ запусков на $400$-ой итерации, а на серии rbg более чем в
$90\%$ запусков на $800$-ой итерации возникают <<хорошие>> задачи
оптимальной рекомбинации, и доля таких запусков увеличивается с
ростом числа итераций.

Таким образом, можно ожидать, что  в ходе работы ГА с элитарной
оптимальной рекомбинацией при решении задачи~$1|s_{vu}|C_{\max}$ в
большинстве случаев будут возникать полиномиально разрешимые
индивидуальные задачи оптимальной рекомбинации, и, кроме того,
размерность области их допустимых решений будет уменьшаться с
ростом числа итераций.

\section{Заключение}\label{sec:Conclusion}
\sloppy
\par
В настоящей  статье разработан генетический алгоритм с
использованием оператора оптимальной рекомбинации, предложенного
в~\cite{ErKo12}. В проведенном эксперименте на каждом из
рассмотренных тестовых примеров из библиотеки TSPLIB генетический
алгоритм находил оптимальное решение более чем в 30\% испытаний. В
ходе работы генетического алгоритма в большинстве случаев
возникали индивидуальные задачи оптимальной рекомбинации с не
более чем~$|V|$ допустимыми решениями и мощность множества их
допустимых решений уменьшалась с ростом числа итераций ввиду
снижения разнообразия популяции.

Универсальность рассмотренного алгоритма оптимальной рекомбинации
состоит в том, что он может использоваться не только при целевой
функции, минимизирующей общий момент завершения выполнения работ,
но и при других критериях (см. примеры в \cite{TKSH, HGE,YI}). 
Только в таком случае его вычислительная сложность будет зависеть
от трудоемкости вычисления целевой функции для заданной
перестановки из области допустимых решений оптимальной
рекомбинации. В связи с этим, для дальнейшего анализа представляет
интерес экспериментальное исследование обобщений предложенного
оператора оптимальной рекомбинации  в составе генетических
алгоритмов для различных задач составления расписаний с
переналадками и его сравнение с другими операторами кроссинговера.
\\

Исследование выполнено при финансовой поддержке РФФИ
(проект~12–01–00122).

\renewcommand{\refname}{ЛИТЕРАТУРА}

\end{document}